# Using Dissortative Mating Genetic Algorithms to Track the Extrema of Dynamic Deceptive Functions

C. M. Fernandes[1,2], J.J. Merelo[2] and A.C. Rosa[1]
[1]LaSEEB-ISR-IST, Technical Univ. of Lisbon (IST)
{cfernandes,acrosa}@laseeb.org
[2]Department of Architecture and Computer Technology, University of Granada, Spain
*jmerelo@geneura.ugr.es*

**Abstract.** Traditional Genetic Algorithms (GAs) mating schemes select individuals for crossover independently of their genotypic or phenotypic similarities. In Nature, this behaviour is known as random mating. However, non-random schemes − in which individuals mate according to their kinship or likeness − are more common in natural systems. Previous studies indicate that, when applied to GAs, negative assortative mating (a specific type of non-random mating, also known as dissortative mating) may improve their performance (on both speed and reliability) in a wide range of problems. Dissortative mating maintains the genetic diversity at a higher level during the run, and that fact is frequently observed as an explanation for dissortative GAs ability to escape local optima traps. Dynamic problems, due to their specificities, demand special care when tuning a GA, because diversity plays an even more crucial role than it does when tackling static ones. This paper investigates the behaviour of dissortative mating GAs, namely the recently proposed Adaptive Dissortative Mating GA (ADMGA), on dynamic trap functions. ADMGA selects parents according to their Hamming distance, via a self-adjustable threshold value. The method, by keeping population diversity during the run, provides an effective means to deal with dynamic problems. Tests conducted with deceptive and nearly deceptive trap functions indicate that ADMGA is able to outperform other GAs, some specifically designed for tracking moving extrema, on a wide range of tests, being particularly effective when speed of change is not very fast. When comparing the algorithm to a previously proposed dissortative GA, results show that performance is equivalent on the majority of the experiments, but ADMGA performs better when solving the hardest instances of the test set.

## 1. Introduction

Optimization in dynamic environments adds a level of complexity above those already present in static optimization. A problem is said to be dynamic when there is a change in the fitness function, problem instance or restrictions, thus making the optimum change as well. When changes occur, solutions already found may be no longer valuable and the process must engage in a new search effort. When using evolutionary [4] and other bio-inspired algorithms to tackle this kind of problems, the crucial and delicate equilibrium needed between exploration and exploitation in static environments becomes even more important and complex to deal with. Other difficulties arise when changes in the environment are not easily detectable, making reactions to those same changes hard to implement. In addition, even if the change is detectable, it is hard to decide if it is better to restart the population or continue the search with the same population after a shift in the environment. That decision may depend on the severity of the changes, or even on their whole dynamics (defined by severity, speed, periodicity, amongst other traits [13] [2]).

Another issue of extreme importance that arises when designing algorithms to solve Dynamic Optimization Problems (DOPs) is the wide spectrum that the environmental dynamics may cover. As it happens in static problems, *No-Free-Lunch* Theorems [70] are incompatible with the project of designing an algorithm that outperforms any other method when tracking moving optima. Whereas in static optimization the difficulties for Evolutionary Computation [4] arise from the fact that the infinite set of search landscapes includes those without some sort of structure that can be exploited, DOPs pose additional difficulties due to the dynamics of the fitness landscape. If the number of environments or the period between changes varies unboundedly, then, on average, no other method is better than random restarts after a change [70]. However research on evolutionary algorithms for DOPs is not doomed to be an aimless task because the goal is not solving all possible problems, but only a narrower subset focused on a particular domain of interest (or under certain bounds).

With these limitations in mind, it may be stated that evolutionary algorithms hold self-adaptive traits that makes then a suitable candidate to solve dynamic problems. However, if the algorithm converges, its ability to acquire the new optimum after a change is reduced. Genetic diversity – that is, the genotypic variety present in a population at a specific moment (generation) – must thus be maintained or least introduced at regular intervals. Mutation (or other method to maintain or introduce genetic diversity in the population) is essential and, in some cases, increasing mutation above the optimal value needed to solve the static problem is enough to track the optima. On the other hand, high rates have a destructive effect [36] − and it is not possible to rely on reactive macro-mutation because changes (or their severity) are not always detectable. Therefore, other methods are required to deal with the complexity of dynamic landscapes in general. Due to the large number of published proposals on evolutionary algorithms for DOPs, a classification of techniques is needed in order to distinguish major approaches and identify their preferable targets. A possible set of categories to classify evolutionary algorithms designed to solve DOPs was proposed by Branke in [12].

Branke's classification of Evolutionary Algorithms for DOPs divided them into three classes: *reaction to changes*, *memory* and *diversity maintenance*. The algorithms that hold some kind of mechanism that reacts to changes and introduce diversity in population are said to belong to the first class. However, these schemes, unless some self-adaptive behaviour is incorporated, assume that changes are detectable. *Memory schemes* may also need change detection mechanisms in order to recover previous solutions. Memory is particularly useful when the dynamics of change is circular, that is, the shape of the fitness landscape repeats from time to time. When the changes lack periodicity, memory schemes lose some of their effectiveness.

A more general way of keeping track of changing solutions is maintaining diversity throughout the run. Schemes that constantly introduce diversity or somehow delay its loss may reduce convergence speed, but when solving DOPs it is sometimes more important to just keep close to the optimum than to completely converge to it. These algorithms are more general and may be applied to a wider range of situations, without being dependent on whether or not changes are detectable or on some specific dynamics of the landscape. Mating is one of the stages where modifications can be made in order to provide means to keep diversity. The investigation reported in this paper deals with *non-random mating* techniques for Genetic Algorithms (GAs) and their efficiency in dynamic environments.

Although genetic and other evolutionary algorithms usually engage in what is called *random mating* [60], [61] − i,e., individuals mate only according to its fitness and not depending on their genotypic or phenotypic similarities −, in Nature non-random mating, which encloses different kinds of strategies based on kinship or likeness of the agents involved in the reproduction game, seems to be the rule amongst sexual organisms. Humans, for instance, mate preferentially outside their family tree, using a non-random mating scheme known as *outbreeding*, which has its opposite in *inbreeding*, a selection strategy in which individuals mate preferentially with their relatives [60] [61]. It is often stated that inbreeding decreases the genetic diversity in a population while outbreeding increases that same diversity [61]. In addition, inbreeding increases the normal rate of a harmful allele present in the family. If inbreeding is extensive and intensive, homozygozity will increase in frequency and the family experiences a growth in the genetic load (measure of all of the harmful recessive alleles in a population or family line) of the harmful allele.

Assortative mating is another kind of non-random strategy, in which crossover between similar or dissimilar is restricted. When similar individuals mate more often than expected by chance, we are in presence of positive assortative mating (or assortative mating in the strict sense). When dissimilar individuals mate more often, the scheme is called negative assortative mating (or dissortative mating). Positive assortative mating results in an average increase in homozygozity and in an increase in population variance [60]. However, this does not mean that genetic diversity is increasing. In fact, this type of mating may result in highly distinct clusters of similar genotypes, thus playing a crucial role when speciation without geographic barriers occurs (sympatric speciation [65]). Dissortative mating, on the other hand, has the primary consequence of a progressive increase in the frequency of heterozygous genotypes; more diversity is a direct consequence of these changes in the genotype frequencies.

Evidences show that mating, besides very unlikely to be random in Nature, may have the potential to act as an evolutionary agent, although its effects are complex and hard to model. Nevertheless, models presented by Jaffe [41] and Ochoa *et al.* [54] shed some light into the subject, and gave empirical support to the hypothesis that mating is not likely to be random in Nature and that assortative and dissortative mating may produce higher survival rates among individuals evolving in, respectively, static and dynamic environments. While in dynamic landscapes genetic variability is fundamental to quick responses to changes, diversity in static environments is less important. In fact, organisms move towards an optimal degree of variability that depends on the environment, via some mating scheme. Environment itself appears to guide the evolution of mating strategies [54].

These issues naturally lead research on Evolutionary Computation toward mating schemes that implicitly or explicitly holds some kind of genotypic or phenotypic restriction. This work studies the effects of dissortative mating on the performance of GAs solving dynamic trap functions. We are particularly interested in algorithm recently proposed by Fernandes *et al*. [28], [31], the *Adaptive Dissortative Mating GA* (ADMGA), and on whether (and in which circumstances) it can improve GAs performance on dynamic optimization. For that purpose, a DOP generator [72] was applied to static trap functions tuned into the regions of nearly deceptiveness and deceptiveness. The results were compared to those attained by other GAs, including a preceding dissortative mating GA (described in the Section 3 along with some of the more relevant work in the field) and two GAs previously applied to dynamic environments. Dissortative mating strategies, when compared to the other GAs in the test set, proved to be more effective when solving dynamic trap functions.

The paper is structured as follows. The next section describes some of the published studies on bio-inspired algorithms for dynamic problems. Section 3 deals with previously proposed evolutionary algorithms holding non-traditional selection methods that may be classified as non-random. Section 4 describes ADMGA. Section 5 describes the experimental setup and Section 6 presents and discusses the results. Finally, Section 7 concludes the paper and outlines possible lines of work to tread in the future.

## 2. Bio-inspired Algorithms for Dynamic Optimization

Besides all the complexity inherent to DOPs, traditional evolutionary algorithms come across some difficulties when solving dynamic problems due to their specific nature (for a detailed survey see [12]). Namely, if the first convergence stage reduces population diversity, then the algorithm may not be able to react to changes. One standard approach to deal with it is to regard each change as the arrival of a new optimization problem that has to be solved from scratch. However, this simple approach is often inefficient when compared to other possible strategies, because solving a problem without reusing information from the past might be time consuming, a change might not be identifiable directly, or the solution to the new problem should not differ too much from the solution of the old problem. Thus, it is sometimes better to have an algorithm that is capable of continuously adapting the solution to a changing environment, reusing the information gained in the past.

The problem here can be stated as seeking an appropriate balance between two contradictory characters of the search procedure, those between the exploring and exploiting nature of the algorithm. Over the past few years, a number of authors have addressed the problem of convergence and subsequent loss of adaptability when solving dynamic problems with evolutionary and other bio-inspired algorithms. The following paragraphs describe some of them and, whenever it is possible, the techniques are classified within the three categories defined by Branke [12]: reaction to changes, memory and diversity maintenance.

Cobb's *Hypermutation* [14] is a technique that may be classified within the first category. Hypermutation deals with changing environments by increasing significantly the mutation rate when a change is detected. However, this algorithm is far from being effective under all circumstance, because, as already stated, changes are not always detectable and small changes in the environment may be easily tracked without the destructive effect of such high mutation rates. *Variable Local Search* [68] [69] is a similar algorithm, in which the mutation rate is gradually increased after a change has been detected. Other methods, like those proposed by Bierwirth and Mattfeld [9] and Lin et al. [45], create a new population after a change using the solutions of the previous population as a seed.

Another kind of approach is to supply the algorithm with some sort of memory that stores good partial solutions in order to reuse them later. This can be advantageous in cases where the environment is changing periodically and repeated situations occur. On the other hand, they can be counterproductive if the environment changes dramatically with open-ended novelty. Memory may be provided in two general ways: *implicitly* by using redundant representations, or *explicitly* by introducing an extra memory and formulating strategies to deposit and retrieve solutions later. The major approach to implicit memory and redundant representation is multiploidy [35] [43], [53] [67]. Then again, while redundant representations allow the evolutionary algorithm to implicitly store some useful information during the run, it is not clear that the algorithm actually uses this memory in an efficient way.

As an alternative, some approaches use an explicit memory in which specific information is stored and reintroduced into the population at later generations, as in [47]. Branke [11] compared a number of replacement strategies for inserting new individuals into a memory stressing the importance of diversity for memory-based approaches. Other approaches based on the concept of memory explicit memory were proposed by Mori *et al*. in [51] and by Trojanowski and Michalewicz in [66]. Recently, Barlow and Smith [8] proposed a memory enhanced evolutionary algorithm for dynamic scheduling problems, and Richter and Yang [58] proposed a memory scheme based on abstraction, i.e., the algorithm does not store good solutions, but instead it keeps their approximate location in the search space.

Unlike the reactive algorithms and some of the memory schemes described above, diversity maintenance techniques do not (in general) depend on whether a change is detectable or not. In addition, their performance is less dependent on the periodicity of the changes than memory-based evolutionary algorithms. A well-known example of a diversity maintenance technique dates back from 1992: the *Random Immigrants GA* (RIGA) [37]. RIGA maintains diversity by introducing $r_r$ random solutions in the population in each generation. This guarantees that brand new genetic material enters the population in every time step, thus avoiding the convergence of the whole population to a narrow region of the search space. RIGA is a kind of standard algorithm for dynamic environments and it appears very often in experimental studies in order to evaluate the performance of new proposals. However, its behaviour is affected by the parameter $r_r$, and, according to the experiments conducted for this paper's investigations, it is not clear that RIGA improves GAs performance on DOPs, at least when solving dynamic deceptive functions (see Section 6). The following algorithms may also be classified in the same category (diversity maintenance) as RIGA.

In [64], a RIGA associated with the Bak-Sneppen model [6] is presented and tested on DOPs: the *Self-Organized Random Immigrants Genetic Algorithm* (SORIGA). The Bak-Sneppen model is known to have Self-Organized Criticality properties, a phenomenon that was detected in 1987 by Bak *et al*. [5], and which is characterized by displaying scale invariant behaviour. When associated with evolutionary algorithms it may periodically insert large amounts of new material in the population or completely reorganize a solution to a problem. For these reasons, researchers soon adopted it in order to provide new means to control parameter values or maintain population diversity. The research field on DOPs was a logical following step. Besides SORIGA, another approach has been recently proposed by Fernandes *et al.* [29], in which the Sand Pile model [5] – also a Self-Organized Criticallity model − is attached to a GA in order to solve dynamic problems. The Sand Pile works together with mutation with the purpose of modeling an on-line behaviour, characterized by small and medium mutation rates punctuated by dramatic reconfigurations of the population. The resulting algorithm appears to self-adapt to the frequency and extent of the changes, that is, to the dynamics of the problem.

Another interesting approach is the *co-evolutionary agent based model of genotype editing* [40], which uses several genetic editing characteristics that are gleaned from the RNA editing system as observed in several organisms. Their results outperformed traditional GAs via obtaining greater phenotypic plasticity.

Finally, Laredo et al. [42] proposed a peer-to-peer GA that maintains diversity at a higher level due to its niching properties. The reported results show that the algorithm is capable of outperforming other GAs on dynamic functions.

In the past decade, relevant work is arising from the Estimation of Distribution Algorithms (EDAs) [46] research field. Research on EDAs − a class of evolutionary algorithms where a probability distribution vector replaces an explicit representation of the population – has experienced a continuous and consistent growth in the last decade. However, only recently the DOP issue has started to raise a strong interest on EDAs' researchers. For instance, the *Population Based Incremental Learning* (PBIL) algorithm [7] - one of the first EDAs - is used by Yang and Yao [72] to solve three different dynamic problems, and several versions of the algorithm are compared with GAs and RIGAs on those problems. In [73], Yang works on a simple EDA and proposes the *Univariate Marginal Distribution Algorithm* (UMDA) [52] with memory and the experiments show that the scheme is efficient in dynamic environments.

More recently, Lima *et al*. [44] investigated the incorporation of restricted tournament replacement (RTR) in the extended compact genetic algorithm (ECGA) [38] for solving problems with non-stationary optima. (RTR is a simple yet efficient niching method used to maintain diversity in a population of individuals.) The results confirmed the assumptions on the efficiency of the method, which proves that niching methods are efficient for keeping diversity.

In [27], Fernandes *et al.* presented the *Binary Ant Algorithm*, which is based on ant algorithms [10], and takes advantage of their ability to solve combinatorial DOPs to generalize them to binary codifications. However, this method may be also regarded as a type of EDA, because the *Binary Ant Algorithm* creates the possible solutions to a problem via transition probability vectors. Later, and following this line of work, Fernandes *et al*. [30] proposed a new update strategy for UMDA, also based on ant algorithms, that aims at maintaining the diversity of the algorithm at a higher level. The new strategy outperformed the standard UMDA and other variations of the algorithm on the majority of the tests with dynamic problems.

We will move on now to the next section, which describes evolutionary algorithms that hold non-random mating strategies. Some of those strategies lead to genetic diversity maintenance, thus making the algorithms good candidates to be applied on dynamic optimization. We are interested in the most robust and context-independent techniques − diversity maintenance techniques − to solve DOPs and, due to their characteristics, non-random mating algorithms are worthwhile exploring as robust moving optimum trackers.

### 3. Non-random Mating in Evolutionary Computation

This section describes evolutionary algorithms with non-random mating schemes. A special emphasis is given to those that, to the extent of our knowledge, were seminal in their line of work, and to those that preceded (or that are, at some level, related to) ADMGA.

A GA with incest prohibition is described in [16], in which individuals with a certain degree of kinship are not allowed to crossover. This policy does not completely restrict mating between similar chromosomes, but it decreases its frequency since related individuals tend to share a large amount of common alleles. The algorithm has been shown to decrease optimal mutation rates. Since outbreeding is supposed to keep populations' diversity at a higher level, reducing the need for mutation to introduce novelty into converging populations, this outcome is expected. Later, Fernandes *et al.* [23] combined the outbreeding strategy proposed in [16] with the GA with varying population size [3] to create the *non-incest Genetic Algorithm with Varying Population Size* (niGAVaPS). Tests made with the algorithm ranging through different degrees of incest prohibition showed improvements in the capability of escaping local optima when the individuals are not allowed to mate with their parents and siblings. However, these algorithms depend much on the degree of parenthood that is defined as a threshold, and an efficient tuning of that parameter is not a straightforward task.

Another (similar) way of improving evolutionary algorithms' performance by maintaining diversity is engaging in dissortative mating schemes. Back in 1984, Mauldin [50] proposed a method to avoid the coexistence of similar individuals in the population based on a Hamming distance restriction. CHC [21] [22] − which stands for *Cross generational elitist selection, Heterogeneous Recombination and Cataclysmic Mutation* − is in some way a descendent of Mauldin's method. CHC is a variation of the standard GA that holds a simple dissortative mechanism. Although the title in [22] may suggest that CHC is an outbreeding GA, a closer look reveal that the algorithm uses a dissortative mating strategy in order to prevent premature convergence. CHC uses no mutation in the classical sense of the concept, but instead it goes through a process of macro-mutation (high rates) when the best fitness of the population does not change after a certain number of generations. Diversity is assured by a highly disruptive crossover operator, the Half Uniform Crossover (HUX) [22], and by a reproduction restriction that assures that selected pairs of chromosomes will not generate offspring unless their Hamming Distance is above a certain threshold. If none of the paired chromosomes in a generation is able to generate offspring, then the threshold is decremented. When the threshold reaches 1 and the algorithm is not able to generate new chromosomes (that is, the population has converged to a local optima), then a cataclysmic mutation is applied by replacing the entire population, except the best chromosome, with mutated copies of that individual.

Inspired, to a certain extent, by CHC, the *Self-Regulated Evolutionary Algorithm* (SRPEA) was proposed by Fernandes and Rosa [26] as an algorithm with a dynamic on-the-fly variation of the population size. Like CHC, selected individuals recombine their genotypes and generate offspring only if their Hamming distance is above a threshold value. That value changes over time, depending on the number of newborn individuals and deaths in each generation. Individuals die (that is, are removed from the population) only when their lifetime (which is set to specific value in the beginning of the search depending on the individual's fitness) reaches zero, which means that parents and children may belong to the same population. An empirical study demonstrated that the algorithm self-regulates its population size: there are neither uncontrolled demographic explosions nor quasi-extinction long stages.

Another possible way of inserting assortative or dissortative mating into a GA is described in [24]. The *Assortative Mating GA* (AMGA) selects, in each recombination event, one parent, by any method. Then, it selects a pool of $n$ individuals. After computing the similarity between the first parent and the $n$ individuals, the second parent is selected according to the type of mating in progress. If the algorithm is the *positive* AMGA (pAMGA), then the individual more similar to the first parent is chosen. The *negative* AMGA (nAMGA) selects the individual less similar to the first parent. The size of the pool controls the intensity of mating restriction. Experiments with the algorithm solving a vector quantization problem showed that positive assortative and standard GA performed similarly, while nAMGA outperformed both [24]. In [25], the algorithm was combined with a varying population size mechanism and compared with a standard GA. The negative assortative mating (or dissortative mating) strategy has proven to be more able in escaping local optima traps. Although nAMGA results are interesting, the size of the pool is critical for its performance. ADMGA was a step towards a dissortative mating algorithm less dependent on parameter tuning. In that sense, and since ADMGA, unlike nAMGA, does not add any parameter to the already complex structure of the traditional evolutionary algorithms, the aim was achieved.

An idea related with AMGA was carried out by Ochoa *et al*. [55] on dynamic environments. Assortative GAs (very similar to AMGA) were used to solve a dynamic knapsack problem. Dissortative mating was more able to track solutions, standard GA often failed to track them, and the assortative GA had the worst performance. The authors also discuss the optimal mutation rate for different strategies, concluding that the optimal rate increases when the strategy goes from dissortative to assortative. These results were predictable: dissortative mating is supposed to keep the population's diversity at a higher level, reducing the amount of mutation needed in order to prevent premature convergence. In this line of work, there is a study by Ochoa *et al*. on the error threshold of replication in GAs with different mating strategies [56]. The error threshold is a critical mutation rate beyond which structures obtained by an evolutionary process are destroyed more frequently than selection can reproduce them. By evolving a GA on different landscapes, the authors conclude that recombination shifts the error threshold toward lower values. In addition, assosrtativity overcomes this effect by increasing the error threshold, while dissortative pushes the error into lower values. The study aims at shedding some light into the relation between mutation rates and mating strategies in evolutionary algorithms.

Another approach is found in [31], where an assortative mating strategy is used to implement a local search genetic algorithm. The approach is consistent with the fact that crossover is the main mechanism of a GA generating local search, and assortative mating, by its own characteristics (it favours crossover between similar individuals), tends to increase the strength of exploitation, thus leading to a more intensive local search. On the other hand, Garcia-Martinez *et al.* [33] intro-duced a real-coded GA with dissortative mating. The authors show that the inclusion of that mating strategy increases the performance of the algorithm on a set of proposed problems. In addition, empirical analysis indicates that the merits of dissortative mating are clearer with lower values of $\alpha$ parameter of the PBX-$\alpha$ crossover [48]. This observation is closely related with the optimal mutation rate issue described above, since $\alpha$ determines the spread of the probability distribution used to create offspring with PBX-$\alpha$. This way, parameter $\alpha$ acts as genetic diversity controller, with higher values leading to GAs with higher exploratory capabilities, as it happens with mutation rate values. Therefore, if dissortative mating is expected to decrease optimal mutation rates, optimal values of $\alpha$ may also be dependent on the mating strategy chosen for the GA, being lower when dissimilar individuals have more chance to generate offspring.

Besides de above-referred techniques, which are directly related to ADMGA and to the line of work that conduced to it, a large number of other GAs with non-random mating are found in the Evolutionary Computation literature. The following paragraphs briefly describe some of them in chronological order.

Hillis [39] described a co-evolutionary computation paradigm with assortative mating applied to a sorting network problem. The author does not provide results comparing the proposed strategy and random mating but it states that the choice on assortative mating was inspired by some problem characteristics rather than genetic diversity concerns.

Ronald [59] introduced the concept of seduction in GAs, which consists in selecting the second parent according to the preferences of the first parent. After the first chromosome involved in a recombination event is selected, all other individuals in the population are provided with a secondary fitness according to certain rules that reflects the preferences of the first parent. Then, the second parent is chosen according to the secondary fitness.

De *et al.* [18] proposed genotypic and phenotypic assortative mating. The schemes are compared with standard GA and CHC on some well-known test functions and on the problem of selecting the optimal set of weights in a multilayer *perceptron*. Phenotypic assortative mating revealed to be the best strategy, outperforming standard GA and CHC on the range of proposed problems.

Matsui [49] incorporated dissortative mating within the tournament selection strategy. After the first parent is selected, the second parent is chosen according to a function that depends on the individual fitness and the Hamming distance to the first parent (all individuals in the population are inspected in order to determine the distance to the first parent). In addition, the author incorporates a family-based selection mechanism that, by applying selection and replacement at family level (two parents and two offspring), maintains the genetic diversity of the population. The algorithm was applied to a dynamic knapsack problem.

Dolin *et al*. [20] proposed the Complementary Phenotype Selection for the recombination procedure in Genetic Programming [62]. The scheme, that may be easily extended to other evolutionary algorithms, selects the first parent (mother) using a typical selection method and then selects a father whose strengths complement the mother's weaknesses. This approach is relevant only in Evolutionary Computation domains that make use of "fitness cases". This means that each individual in the population is tested over some finite, representative set of problem examples, and each individual's fitness is based on performance over this set. Therefore, "strength" and "weakness" refers to the scores of each individual on the set of problems.

Ting *et al*. [63] introduced the Tabu Genetic Algorithm (TGA). TGA combines the characteristics of GAs and Tabu Search [34], by incorporating a taboo list in a traditional GA that prevents inbreeding and maintains genetic diversity. TGA also uses an aspiration criterion [63] in order to allow some crossovers even if they violate the taboo. Since incest prevention efficiency is sensitive to mutation rate, the authors include a self-adaptive mutation in TGA. The process is somehow similar to the cataclysmic mutation that occurs in CHC, since mutation in TGA occurs in presence of a deadlock situation, that is, when the genetic diversity of the population as decreased down to a level were allowed recombination is almost or even impossible to occur.

Finally, Wagner and Affenzeller [71] introduced the SexualGA, which simulates sexual selection within the frame of a GA and uses two different selection schemes in the same population.

Unlike most of the previous methods, the algorithm (ADMGA) proposed in this paper for dynamic optimization, self-regulates the intensity of dissortative mating (the closer to ADMGA is CHC, but the later uses reinicialization procedures). ADMGA, as shown in [28], maintains genetic diversity at a higher level. It is this ability and the simplicity (self-regulation) of the algorithm that we aim at exploring in order to solve DOPs. As stated, diversity maintenance techniques are, in general, more robust to the wide range of dynamics that a specific dynamic problem may cover. These algorithms do not depend on either changes are detectable or not (unlike algorithms that react to changes), do not rely on the periodicity of changes (unlike memory schemes) and may work well on both small and dramatic changes in the environment (unlike macro-mutation techniques). ADMGA holds these characteristics and arises as a possible technique that can cover a wide spectrum of dynamics with stable efficiency. In addition, as stated above, ADMGA does not increase the parameter space of traditional GAs and it self-regulates its behaviour.

## 4. ADMGA

ADMGA [28], proposed by Fernandes *et al*. in [28], is a self-regulated dissortative mating evolutionary algorithm that incorporates an adaptive Hamming distance mating restriction that tends to relax as the search process advances, while it may also be occasionally reinforced. After two parents are selected, crossover only occurs if the Hamming distance between them is found to be above a threshold value. If not, the recombination event is considered as "failed" and another pair of individuals is selected until $N/2$ pairs have tried to recombine ($N$ is the population size). After the reproduction cycle is completed, a new population is created by selecting the best $N$ members amongst the parents and newly generated offspring (ADMGA is a steady-state GA) and the amount of successful and failed crossovers is compared. Then, the threshold is incremented or decremented, depending on the number of successful and failed events (see pseudo-code). This way, the population diversity controls − indirectly − the threshold value. When diversity decreases, threshold tends to be decremented because the frequency of unsuccessful mating will necessarily increase. However, mutation introduces variability in the population, resulting in occasional increments of the threshold that moves it away from 0 [28]. ADMGA was tested in a wide range of problems and the results showed that the method is capable of outperforming several other GAs (including CHC) on that particular set of functions [28].

Due to the characteristics of the algorithm, dynamic optimization appeared as a valid following step in the research on the behaviour of ADMGA. Some results were already reported in [31] but additional experiments are needed in order to fully understand the capabilities and performance of this method. For solving DOPs, two modifications were made in the original ADMGA: the first concerns the replacement strategy − while in ADMGA for stationary problems offspring compete with the parents' population for a place in the new population, the algorithm's version for DOPs replaces the worst $N$' parents to provide space for the new $N$' individuals. This option gives rise to a less elitist algorithm, and reduces computational effort when assuming that changes are undetectable, because, if changes are undetectable, all the chromosomes in the population are evaluated (or re-evaluated) in each generation. The lower the number of new individuals inserted in the population, the higher is the redundant effort dedicated to re-evaluate solutions (it is not redundant only when a change occurs).

```
   Algorithm: ADMGA
initialize Population(P) with size(P) = N
evaluate Population(P)
set initial threshold(iT) /* iT←L-1, static; iT←L/4, DOPs*/
threshold(T) ← iT
   while (not termination condition)
     create new individuals P.new
     evaluate new individuals P.new
     if (static problem)
       P ← P+P.new
       remove worst individuals from P until size(P) reaches N
      end if
      if (DOP) replace size(P.new) worst individuals by P.new
    end while

Procedure: create new individuals
matingEvents ← N/2;
successfulMating ← 0;
failedMating ← 0
while (successfulMatings < 1) do
   for (i ← 1 to matingEvents)  do
     select two chromosomes (c1, c2)
     compute Hamming distance H(c1, c2)
     if (H(c1, c2) >= T)
          crossover and mutate
          successfulMating ← successfulMating+1
     end if
     if (H(c1, c2) < T)  failedfulMating ← failedlMating+1
   end for
   if (failedMating > successfulMating)  T ← T-1
   else       T ← T+1
end while
```

The second modification is focused on the initial threshold value. Setting the initial value to $L$-1 – as in [28] − is not suited for dynamic optimization. Although ADMGA self-regulates the threshold in the first generation, according to the problem [28], it creates few new individuals until the threshold reaches a value around $L/2$. If the algorithm passes through an initial stage during which few new chromosomes are created, until it reaches a more stable threshold value, then a prohibitive number of re-evaluations are performed, delaying the algorithm and compromising the first stage of optimization, especially when the changes occur fast. For that reason, the initial threshold is different for DOPs. For this study, the initial value was set to $L/4$.

**5. Experimental Setup**

In order to investigate how ADMGA's behaves on hard dynamic problems, experiments were conducted with trap functions [1] [19] used as subproblems to construct larger problems. A trap function is a piecewise-linear function defined on unitation (the number of ones in a binary string) that has two distinct regions in the search space, one leading to a global optimum and the other leading to the local optimum (see figure 1). Specifically, an order-$l$ trap function can be expressed as:

$$trap(u(\vec{x})) = \begin{cases} \frac{a}{z}(z - u(\vec{x})), & if\ u(\vec{x}) \leq z \\ \frac{b}{l-z}(u(\vec{x}) - z), & otherwise \end{cases} \quad (1)$$

where u($\vec{x}$) is the unitation function, defined as:

$$u(\vec{x}) = u(x_1, \ldots, x_l) = x_1 + \cdots\ x_1 = \sum_{i=0}^{l} x_i \quad (2)$$

and $a$ is the local optimum, $b$ is the global optimum, $l$ is the problem size (order-$l$ trap function) and $z$ is slope-change location separating the attraction basin of the two optima as depicted in figure 1.

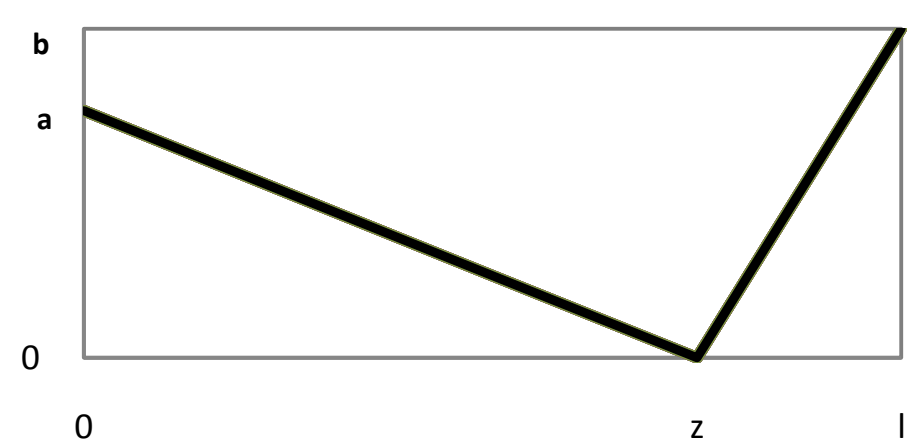

**Fig. 1.** Generalized $l$-order trap function.

Depending on the parameters setting, trap functions may be deceptive or not. Deceptive problems are functions where low-order building-blocks do not combine to form higher order building-blocks. Instead, low-order building-blocks may mislead the search towards local optima, thus challenging GAs search mechanisms. For a trap function to be deceptive, the ratio r between the local ($a$) and global ($b$) optimum must be so that:

$$r \geq \frac{2 - \frac{1}{L-z}}{2 - \frac{1}{z}} \quad (3)$$

In the experiments, order-3, order-4 and order-5 trap functions were designed using the following parameters: a = $l$-1; $b$ = $l$; z = $l$-1. This way, equation 1 may be simplified:

$$F(u(\vec{x})) = \begin{cases} l, & if\ u(\vec{x}) = l \\ l - 1 - u(\vec{x}), & otherwise \end{cases} \quad (4)$$

Please note that with these settings, the ratio r of the order-3 trap function is equal to the threshold, which means that the function lies in the region between deceptive and non-deceptive, while order-4 and 5-order functions are fully deceptive since the condition in equation 3 is satisfied. Under these settings, it is possible to investigate not only how standard GAs and ADMGA performs on dynamic $l$-trap functions, but also to observe how that performance varies when moving from nearly deceptive to fully deceptive search spaces, and when increasing the size of the problem by changing the subprob-lems from order-3 to order-5. For that purpose, *l*-bit decomposable functions were constructed by juxtaposing *m* trap functions and summing the fitness of each sub-function to obtain the total fitness:

$$f(\vec{x}) = \sum_{i=1}^{m} trap(\vec{x}_i) \quad (5)$$

For this study, 30-bit, 40-bit and 50-bit problems were designed by juxtaposing ten order-3, order-4 and order-5 trap functions, respectively. Then, the test environment proposed in [72] was used to create an experimental setup based on the functions described above. Given a stationary problem $f(\boldsymbol{x})$ ($\boldsymbol{x} \in \{0,1\}^L$) where L is the chromosome length, the dynamic environments are constructed by applying a binary mask $\mathbf{M} \in \{0,1\}^L$ to each solution before its evaluation:

$$f(\boldsymbol{x}, t) = f\big(\boldsymbol{x} \text{ XOR } \mathbf{M}(k)\big) \quad (6)$$

where t is the generation index, $k = t/\tau$ ($\tau$ measures generations) is the period index and $f(\boldsymbol{x}, t)$ is the fitness of $\boldsymbol{x}$. M $(k)$ is incrementally generated as follows:

$$\mathbf{M}\,(k) = \mathbf{M}(k-1) \text{ XOR } \mathbf{T}(k) \quad (7)$$

where T$(k)$ is an intermediate binary mask for every period k. This mask T$(k)$ has $\rho \times L$ ones, where $\rho$ is a value between 0 and 1.0 which controls the intensity or severity of change. Please note that $\rho = 0$ corresponds to a stationary problem since T vectors will carry only 0's and no change will occur in the environment. On the other hand, $\rho = 1$ will guarantee the highest degree of change; for instance, if a solution to a problem is a vector of 1's, then the dynamic solution will oscillate between a vector of 1's and a vector of 0's. Therefore, by changing $\rho$ and $\tau$ in the previous set of equations it is possible to control two of the most important features when testing algorithms on DOPs: severity ($\rho$) and speed ($\tau$) of change [13] [2].

Although the speed of change in the above-described model is measured in generations between each change ($\tau$), this paper uses the number of evaluations between each change ($\varepsilon$). Since GAs may have any population size, measuring the speed in terms of generations can lead to unclear reports on the results and their significance. In addition, measuring the speed of change in generations makes it hard to compare GAs with different population size. This approach aims at clarifying the conclusions on the GAs' performance and reducing the ambiguity of the comparative studies. Following these ideas, 12 different scenarios were designed by setting $\varepsilon$ = (2400, 24000, 48000) and $\rho$ = (0.05, 0.3, 0.6, 0.95), for each of the stationary problems. Every run covered 10 periods of change. For each experiment, 30 independent runs were executed with the same 30 random seeds.

With the speed ($\varepsilon$) set to the above values, the algorithms experience the changes when they are at different stages of the search. However, some modifications were made to these settings in order to deal with some issues and difficulties that came up when solving the order-4 and, especially, the 5-order DOP. The details and the reasons for altering the settings are described in the next section.

As for the GAs' parameters, crossover probability, $p_c$, was set to 1.0 for all the algorithms in the test set. On the other hand, mutation rates ($p_m$) covered a wide range of values. Since GAs for dynamic problems rely much more on diversity maintenance than those designed for stationary problems, it is of extreme importance to test several $p_m$ values, others than those usually regarded as standard for static optimization. In addition, some algorithms increase or maintain diversity by other means besides mutation (like RIGAs, for instance). It is thus expected that the optimal mutation rates of those algorithms are different. Finally, since this study is mainly focused on dissortative and assortative mating GAs, and, as stated in Section 3, non-random mating strategies affect the optimal mutation rate values, it is thus mandatory to use different rates when comparing random and non-random mating schemes. Otherwise, the results may be misleading, because an experimental study that considers only one $p_m$ value will certainly bias the results towards one of the mating schemes.

The population size ($N$) also strongly affects the performance of the GAs, not only on static problems, but also on dynamic environments. When the fitness is changing (and when measuring the speed of change in evaluations/time), the population size is of extreme importance. Sometimes, as described in the next section, small populations lead to better results. On the other hand, those same small populations may experiment some difficulties when solving harder instances of the problem (like increasing order from 4 to 5). Although this investigation does not aim at studying scalability or finding the optimal population size for each problem, the above referred issues makes it necessary that a proper research considers different $N$ values, otherwise there is a risk of comparing suboptimal parameter settings and, consequently, getting invalid conclusions.

As for the operators, all the algorithms and experiments were conducted with 2-elitism (the two best solutions are always kept in the population), bit-flip mutation and uniform crossover. Besides $p_c$, $p_m$, $N$ and operators, other parameters are exclusively related to some of the GAs in the test set. Those parameters and their values will be addressed in the following section whenever it is relevant.

In the experiments, ADMGA is compared with a generational GA (GGA) and a Steady-State GA (SSGA) that replaces half of the population in each generation (please remember that ADMGA is a Steady-State algorithm, so it is convenient to introduce a standard GA with those characteristics in the test set). RIGA was introduced in the experiments, and also was SORIGA [64]. Finally, ADMGA is compared with the previously proposed nAMGA, used in [55] to tackle a dynamic knapsack problem with good results. pAMGA was also introduced in the experiments in order to compare the performance of assortative and dissortative mating strategies on this type of functions.

In order to evaluate the algorithms, their performance on the dynamic problems is measured by the mean best-of-generation fitness [72]: the best-of-generation fitness averaged across the number of runs and then averaged over the whole run, as defined in equation 8:

$$\overline{F_{BG}} = \frac{\sum_{i=1}^{G} \frac{\sum_{j=1}^{R} F_{BG_{ij}}}{R}}{G} \quad (8)$$

where $G$ is the number of generations, $R$ is the number of runs (30 in all the experiments) and $F_{BG_{ij}}$ is the best-of-generation fitness of generation i of run j of an algorithm on a specific problem. Sometimes, it is important to average $\overline{F_{BG}}$ values over a certain number of scenarios. In that case, we refer to the metrics as averaged $\overline{F_{BG}}$.

As a final note on the experimental setup we would like to address some conditions and assumptions in which the experiments were conducted and its implications on the behaviour of ADMGA. When evaluating GAs on DOPs it is crucial to avoid comparing algorithms that need different computational efforts between changes. Otherwise, results are misleading. Since this study assumes that changes in the environment are not detectable (this is the most general assumption), all chromosomes are evaluated in each generation, even those that have already been evaluated in a previous generation, as in ADMGA. (ADMGA is a steady-state algorithm, and so parents and children may belong to the same population; for the same reason, SSGA also reevaluates the fraction – half − of the population that has not been replaced by children.) This way, and like generational GAs, ADMGA always performs *N* fitness calculations in each generation but only a fraction of those evaluations are performed on new individuals. Please note that assuming non-detectable changes penalizes ADMGA's performance, because otherwise the algorithm could create more generations between changes (remember that the period between changes in this study is measured by function evaluations). In particular, the fact that ADMGA introduces fewer chromosomes in the population (in each generation) than those that are evaluated should particularly penalize its behaviour on very fast changing environments. This hypothesis is confirmed by the experiments.

## 6. Results and Discussion

Comparing the performance of algorithms on dynamic environments can be more complex than the analysis of results on stationary problems. First, the problem of measuring the efficiency throughout the run − and not only the results or speed of convergence, as in static problems − has to be dealt with. This difficulty is solved, in part, by the mean best-of-generation fitness metrics defined in the previous section, but sometimes a proper evaluation of the algorithm is only attained when visualizing a graph representing the values of the best fitness during the run. This way, it is possible not only to perceive how the algorithm reacts to changes, but also how the algorithms behaves in different periods of the search.

Another difficulty arises from the usual test sets. Defining a wide range of dynamics − by varying either speed or severity of changes − results is a large amount of different problems, even if they are designed using the same stationary function. For instance, in the experiments described in this section there are 3 different static functions and from each one 12 different dynamic scenarios are created, thus making 36 different problems. It is reasonable to expect that no algorithm outperforms all the others in such large test set. Therefore, the comparisons sometimes are made considering sets of scenarios that share some characteristic (for instance, all the scenarios with the same speed of change) so that conclusions may be drawn about the range of applicability of a certain algorithm. In addition, each GA may require different parameter values in order to properly tackle each one of the scenarios. This means that the analysis of the results must be very cautious so that the comparisons are not focused on suboptimal configurations of the algorithms.

These comparisons can be made at different levels: it is possible to evaluate the performance of the algorithms in each scenario, or in each type of dynamics (for instance, the algorithms can be compared in fast environments by averaging the results on those kinds of scenarios). The decisions on the "deepness" of the comparisons must be made case by case, based on the mean best-of-generation fitness values and the best fitness plottings. In the experimental study described in this section, a broad analysis that considers the results on all the scenarios with the same speed of change is always performed before engaging in a comparison scenarios by scenario, together with a statistical analysis of the results. This procedure was chosen (instead of comparing, in a first stage, all the scenarios with the same severity) because the optimal mutation rates are more dependent on speed than severity.

With these issues in mind, we now proceed to the analysis of the results by comparing ADMGA with each type of algorithm in the test set: standard, random immigrants and assortative/dissortative mating GAs.

### 6.1. Standard GAs

Before evaluating and comparing the performance of the GAs it is necessary to determine the size $N$ that leads to better results. Figure 2 shows the mean best-of-generation averaged over all the four scenarios for each speed of change ($\varepsilon$) value (order-3 dynamic problem). The $x$-axis gives the mutation rate values, which ranged from $p_m = 1/(16\times L)$ to $p_m = 4/L$, were $L$ is the problem size ($L = 30$, for the order-3 dynamic problem). The $y$-axis gives the mean best-of-generation values averaged across the four scenarios.

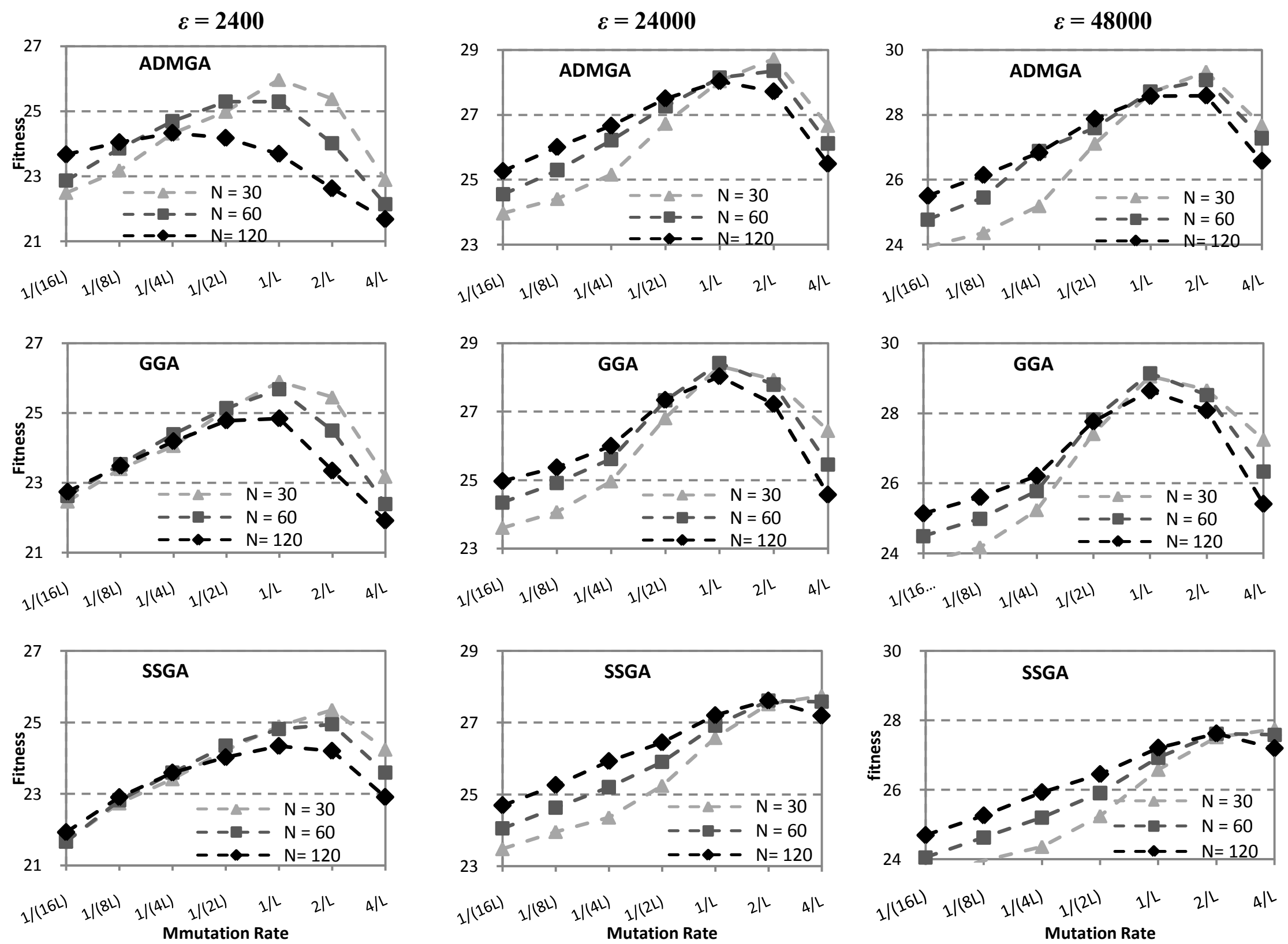

**Fig. 2.** Order-3 dynamic problems: GGA, SSGA and ADMGA performance variation when population size $N$ changes.

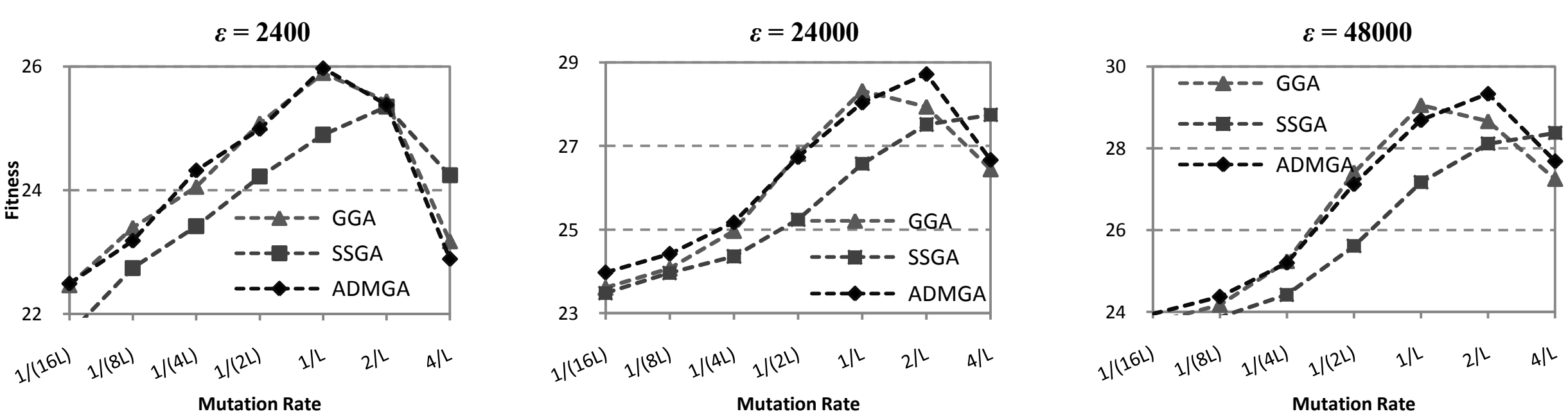

**Fig. 3.** Order-3 dynamic problems: comparing GGA, SSGA and ADMGA performance with population size $N = 30$.

It is clear that small populations perform better and that, in general, $N = 30$ is the best choice for this problem. An overview of the graphics also suggests that ADMGA achieves better fitness values than the other GAs, except on the scenarios with fast speed of change. This outcome is confirmed in figure 3, where the results of each algorithm with $N = 30$ are visualized together. These graphs not only confirm that ADMGA outperforms GGA when $\varepsilon = 24000$ and $\varepsilon = 48000$, but also that the Steady-State GA is the worst algorithm in this test, whatever kind of scenario is considered. Probably, SSGA is affected by the fact that only $N/2$ new chromosomes are introduced in the population in each generation, diminishing the abilities of the algorithm to converge and readapt to changes. On the other hand, ADMGA, which also inserts less than $N$ new chromosomes in the population at each generation, appears to overcome this limitation, except when $\varepsilon$ is low, as expected.

Throughout the entire test set, ADMGA performs worst when facing fast scenarios. The fact that ADMGA is a steady-state algorithm − and so it does not replace the entire population in each generation − is a reasonable explanation for that behaviour. On the other hand, the GA behaves better than most of the algorithms, in most of the order-3, order-4 and 5-order of the scenarios with higher ε. This is probably because ADMGA, although is slower in the beginning of the search due to the kind of replacement, is able to maintain genetic diversity at a higher level [28] and thus it overcomes local optima traps. This characteristic penalizes the algorithm when the fitness function changes quickly, but it helps it to track the optimum when the changes are slow.

The algorithms' results on order-4 trap function were similar to those attained with order-3 dynamic problems. As seen in figure 4, ADMGA's best results are still attained with $N = 30$, although, in this case, $N = 60$ performance is very similar when $\varepsilon = 24000$ and $\varepsilon = 48000$. The reason may be the fact that increasing the order of the problem from 3 to 4 demands more genetic diversity, and that diversity is provided, in the beginning of the search, by larger populations. On the other hand, the dynamic environment typically demands small populations so that there are more generations between each change and an adaptation to change may take fewer evaluations. There must be a balance between the need for larger populations to deal with the stationary problem, and small populations that provide the algorithm with more generations to (re)adapt to changes. Below, when analyzing 5-order problems, we will see that these small populations are not adequate to deal with such a difficult problem. However, for the analysis of the behavior on the order-4 problem, we will use $N = 30$ to compare the algorithms. Figure 5 shows that ADMGA performs again very similarly to GGA on fast environments. Again, like in the test with the order-3 trap, ADMGA is capable of outperforming both standard GAs on slower scenarios, with $p_m = 1/(2\times L)$.

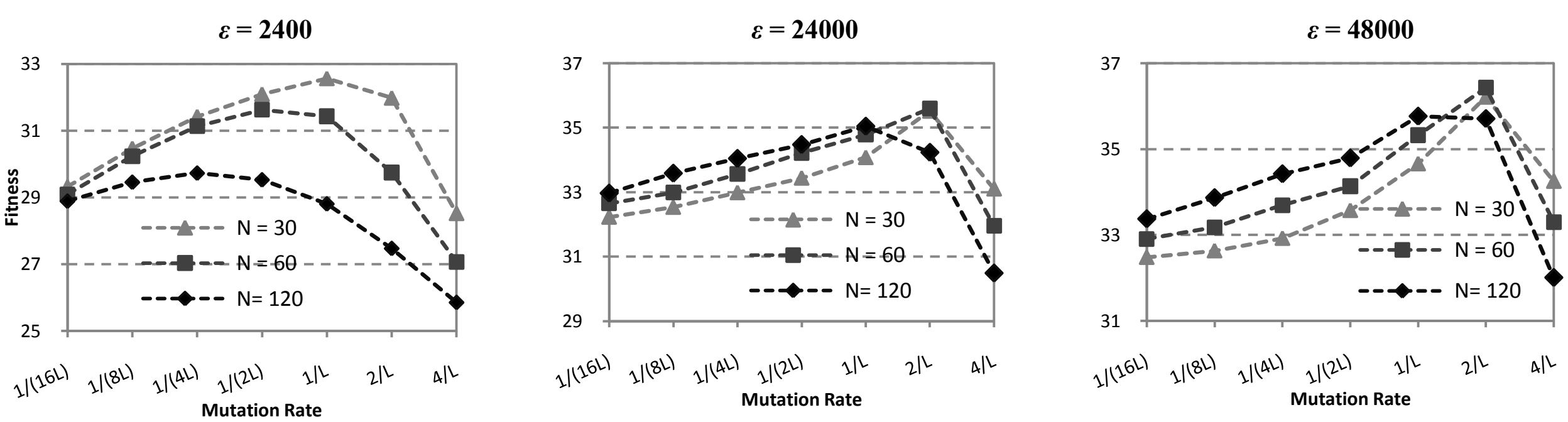

**Fig. 4.** Order-4 dynamic problems: ADMGA performance variation when population size $N$ changes.

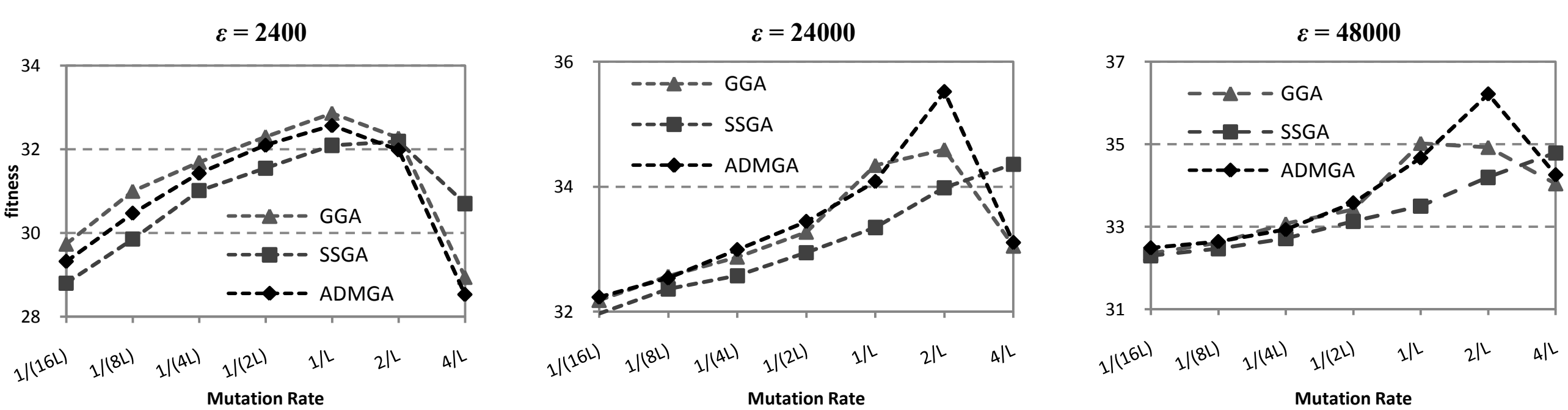

**Fig 5.** Order-4 dynamic problems: comparing GGA, SSGA and ADMGA performance with population size $N = 30$.

In order to evaluate the significance of these results, it is essential to carry out statistical tests. But before that, it is necessary to select which configurations of the algorithms to compare and test. Therefore, by looking at figure 3 and 5, the configuration with the $p_m$ that attained the best results in each kind of scenario was selected. The results of the t-tests are described in table 1. In general, ADMGA performs better than SSGA in all kinds of scenarios. When compared to GGA, ADMGA is capable of outperforming it on almost all scenarios with $\varepsilon = 24000$ and $\varepsilon = 48000$. Nevertheless, an unexpected outcome occurred when comparing the t-test corresponding to order-3 and order-4 problems: when solving order-4,GGA and SSGA are statistically equivalent to ADMGA when $\rho = 0.95$, even when speed is slow. Apparently, increasing the difficulty of the problem caused the efficiency of ADMGA on that particular scenario to decrease, but figure 6 shows that the algorithms are clearly incapable to deal with the changing environments in this case. Unlike the scenario with $\rho = 0.05$, in which both GAs are able to track the optimum, even if GGA is slower, the configuration with $\rho = 0.95$ causes the algorithms to oscillate between suboptimal solutions. ADMGA and GGA do not react properly to the changes and after a shift in the environment, and the search barely escapes the previous optima.

Please note that, due to the characteristics of these functions, if an algorithm converges to the global optimum, a dramatic change in the environment will automatically place the population near the local optimum. For instance, if the algorithms are tracking the solutions of an order-4 trap problem with 10 juxtaposed subproblems, the best solution has fitness 40 and the local optimum has fitness 30. If a GA fully converges to the optimum and then the environment shifts with high $\rho$ values, all the previous solutions will have then a fitness equal to 30. The same goes if only some subproblems converge to the global optimum and others to the local optimum. For instance, if two building-blocks converge to the local optimum (and the rest to the global), then the fitness of the solutions will be only 38. Then, when a radical change occurs, most of the chromosomes are evaluated with fitness equal to 32. If the algorithm is unable to react to those severe changes, the GA will oscillate between values close to 38 and 32. That is what is happening to a certain extent in the order-4 problems configurations with $\rho = 0.95$, as shown in figure 6. The effect is particularly visible with GGA with $N = 30$, which is oscillating between values around fitness 36 and 34.

Increasing the population, as shown in figure 6, apparently does not resolve the difficulties. Therefore, another test was conducted in order to properly compare how the algorithms react to severe changes. For that purpose, $\varepsilon$ was increased to 240000. Some of the results are shown in figure 7. When $\rho = 0.05$, ADMGA still outperforms GGA, tracking the optimum faster and keeping closer to it in most of the periods. When $\rho = 0.95$, ADMGA manages to regain the "contact" with the optimum, although it loses some efficiency from the first to second period of changes. GGA, besides remaining further away from the optimum, still oscillates between suboptimal solutions in some of the periods.

This sections ends up with a final note on the behaviour of the algorithms on fast environments. As seen in figure 8, ADMGA, when tracking small changes ($\rho = 0.05$), is slower in the beginning of the search but as the environments shifts, it is able to keep tracking the optimum, while GGA gets stuck in a local optimum with fitness around 34. Although table 2 indicates that the algorithms' results are statistically equivalent in this particular scenario, the observation of the best-of-generation values throughout the run suggests that ADMGA mean best-of-generation values are being pulled down by the solutions in the beginning of the search. As the search proceeds, ADMGA shows that it is much more able to accomplish the objectives in this particular test. The mean best-of-generation values and consequent statistical analysis are affected by the number of periods of change chosen for this test.

**Table 1.** Statistical comparison of algorithms by paired two-tailed t-tests with 58 degrees of freedom at a 0.05 level significance. The t-test result is shown as + sign when ADMGA is significantly better than GGA or SSGA, - sign when ADMGA is significantly worst, and ~ sign when the algorithms are statistically equivalent. ADMGA with $p_m = 1/L$ when ε = 2400 and $p_m = 2/L$ when ε = 24000 and ε = 48000. GGA with $p_m = 1/L$ for all $\varepsilon$. SSGA with $p_m = 2/L$ when ε = 2400 and $p_m = 4/L$ when ε = 24000 and ε = 48000.

| | | ε = 2400 | | | | ε = 24000 | | | | ε = 48000 | | | |
|---|---|---|---|---|---|---|---|---|---|---|---|---|---|
| | | ρ=0.05 | ρ=0.3 | ρ = 0.6 | ρ = 0.95 | ρ = 0.05 | ρ = 0.3 | ρ = 0.6 | ρ = 0.95 | ρ = 0.05 | ρ = 0.3 | ρ = 0.6 | ρ = 0.95 |
| order-3 | GGA | + | ~ | - | ~ | ~ | + | + | + | ~ | + | + | + |
| | SSGA | + | + | + | ~ | ~ | + | + | + | ~ | + | + | + |
| order-4 | GGA | ~ | - | - | - | + | + | + | ~ | + | + | + | ~ |
| | SSGA | + | + | + | ~ | + | + | + | ~ | + | + | + | ~ |

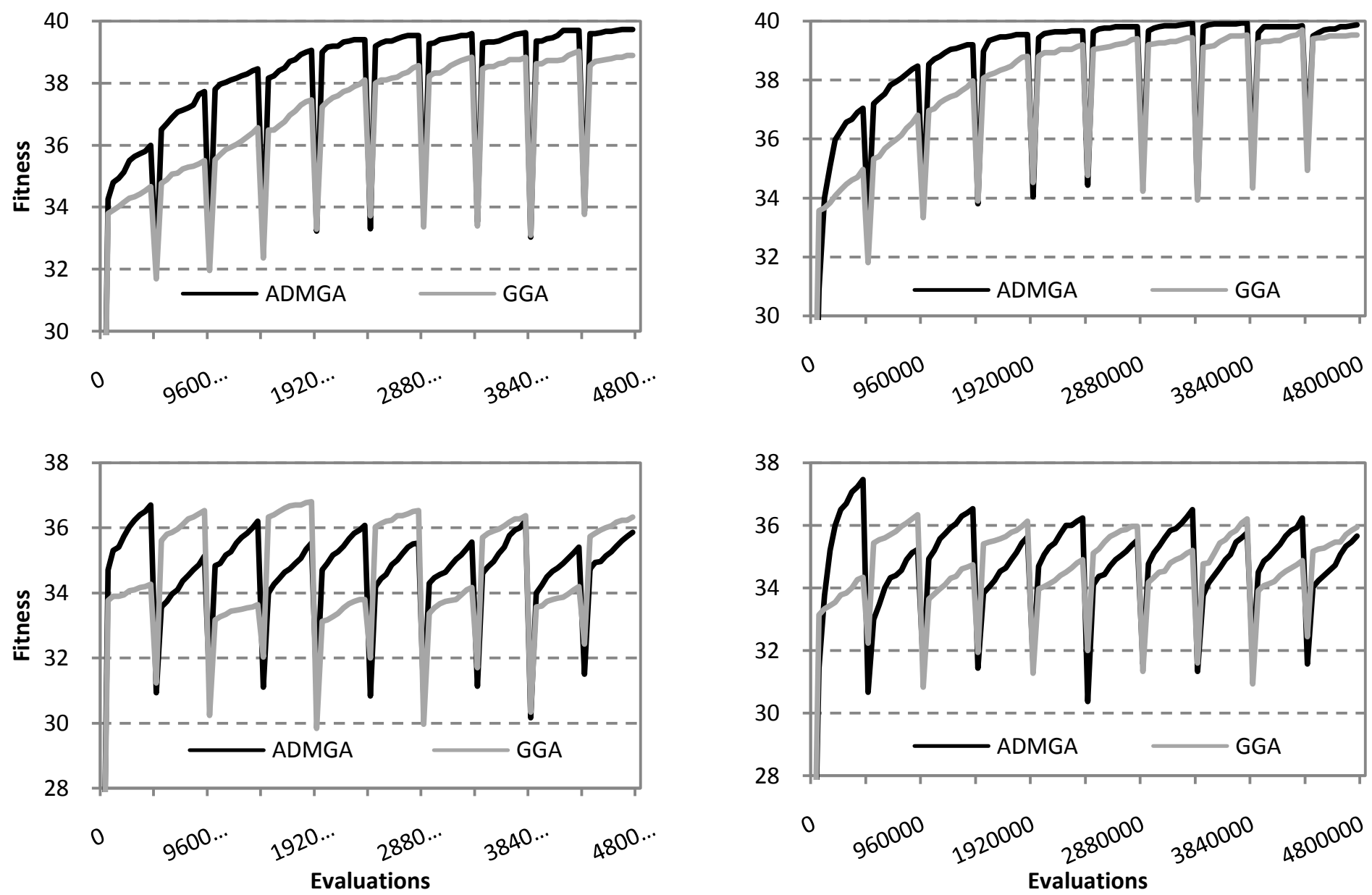

**Fig. 6**. Order-4 dynamic problems. The graphics show the best-of-generation values measured throughout the run for $\rho = 0.05$ (first row) and $\rho = 0.95$ (second row); $\varepsilon = 48000$. GGA ($p_m = 0.025$) and ADMGA ($p_m = 0.05$) with $N = 30$ (left column) and $N = 60$ (right).

When $\rho$ is increased to 0.3, again the same effect is observed: ADMGA is slower and so it takes more periods until it is able to get closer to the optimum than GGA. Finally, when $\rho = 0.95$, the algorithms are again (like in figure 6) not capable of dealing with such severe and fast changes, and the best-of-generation oscillates between two values. The behaviour is more visible in the GGA curves, which clearly oscillate between 33 and 37, meaning that, as stated above, the algorithm is not able to leave the region towards it has converged during the first period of change.

In general, the above results show that ADMGA is capable of outperforming two standard GAs on a nearly deceptive and on a deceptive dynamic trap function, when the speed of change is not very fast. When $\varepsilon$ is low, ADMGA, has some difficulties in converging and reacting to changing optima, but even so its results are statistically equivalent to GGA in most of the scenarios and sometimes it is more able to track the optimum even if it needs more periods to get closer to it. In the next subsection, ADMGA is compared to a traditional and a modified Random Immigrants GA.

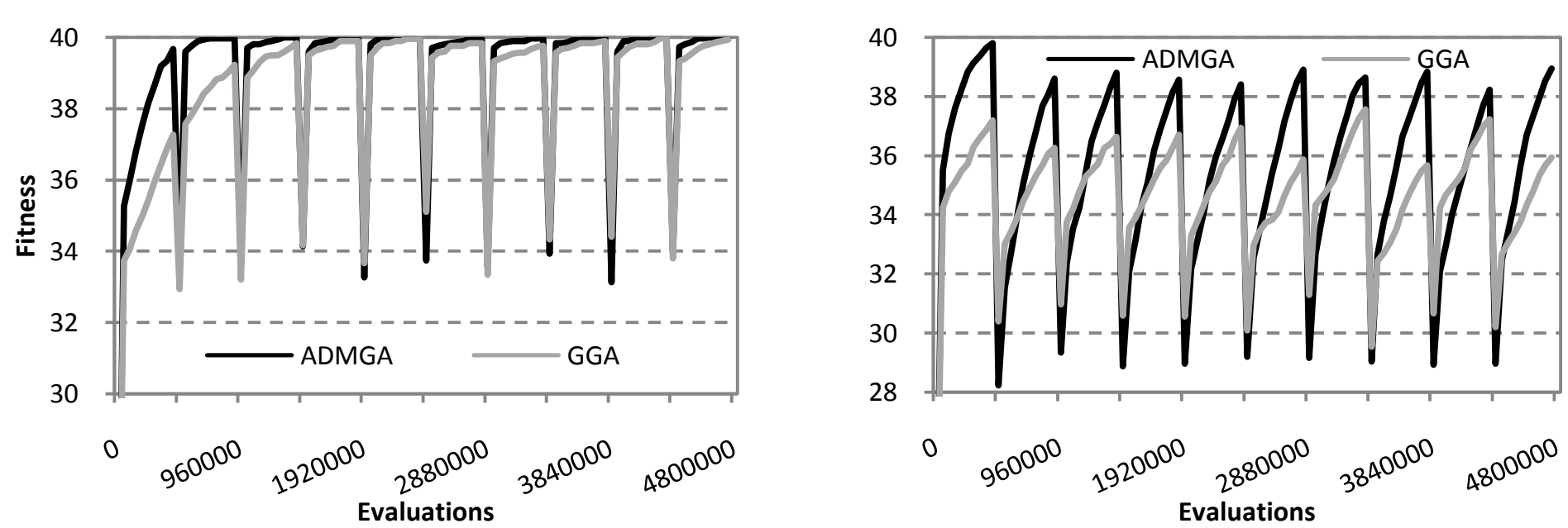

**Fig. 7**. Order-4 dynamic problems. The graphics show the best-of-generation values measured throughout the run for $\rho = 0.05$ (left) and $\rho = 0.95$ (right); $\varepsilon = 240000$. GGA ($p_m = 0.025$) and ADMGA ($p_m = 0.05$) with $N = 30$ are compared.

*6.2 Random Immigrants GAs*

Random immigrants GAs appears very often in evolutionary computation experimental studies on DOPs as a kind of standard algorithm to which new proposals are compared with. For these experiments, two RIGAs were tested: one in which the $r_r$ random immigrants replace the $r_r$ worst chromosomes, and the other in which the random immigrants replace randomly chosen individuals in the population. Both algorithms behaved similarly throughout the test set. The parameter $r_r$ was varied according to the population size $N$ (for $N = 30$, $r_r$ was set to 4). In addition, a more sophisticated random immigrants GA, was also introduced in this experiments: SORIGA (described in Section 2). Reports [64] say

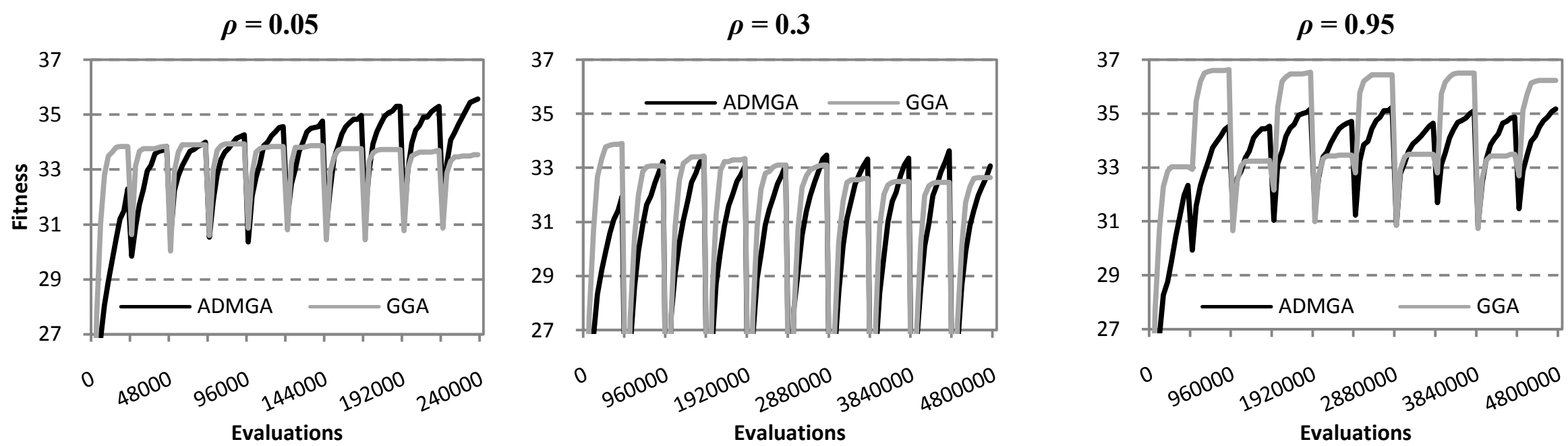

**Fig 8**. Order-4 dynamic problem. The graphics show the best-of-generation values measured throughout the run for different severity values and ε = 2400 (fast changes). GGA ($p_m$ = 0.025) and ADMGA ($p_m$ = 0.025) with $N$ = 30 are compared

that SORIGA was capable of outperforming traditional RIGAs on two deceptive dynamic environments. However, the results of this experimental study show that SORIGA could not attain better results than the standard RIGAs – see figure 9. There are some possible explanations for these results. First, although this study also deals with deceptive functions, those are not the same as the ones used in [64]. Another explanation may reside in the fact that in [64] only a $p_m$ value was considered throughout the entire test set. If, for instance, we check the results attained with that value ($p_m$ = 0.01) in figure 9, then we see that the value lies in the region − between $1/(4\times L)$ and $1/(2\times L)$ − where SORIGA clearly outperforms RIGA.

These results clearly emphasize the need to test the algorithm with different mutation rates. As seen in figure 9 and also in some of the graphs in the previous section, the performance of the algorithms is very dependent on $p_m$. In addition, the optimal mutation rate is different for each GA. In this particular case (figure 9), it is clear that if the experiments only considered $p_m = 1/L$, the conclusions were that the algorithms behave similarly when $\varepsilon$ is bigger than 2400 – see figure 9. On the other hand, if the chosen $p_m$ was lower than $1/L$, then SORIGA would have come out of the experiments as the most effective GA on these dynamic functions. However, considering a wide range of mutation rates, the results show that ADMGA is more effective than both Random Immigrants GAs on the majority of the scenarios.

Table 2 summarizes the results by comparing the best configurations of the algorithms via t-tests. The outcome of the statistical analysis confirms the previous assumptions: ADMGA outperforms both algorithms in most of the scenarios, namely when changes are not very fast.

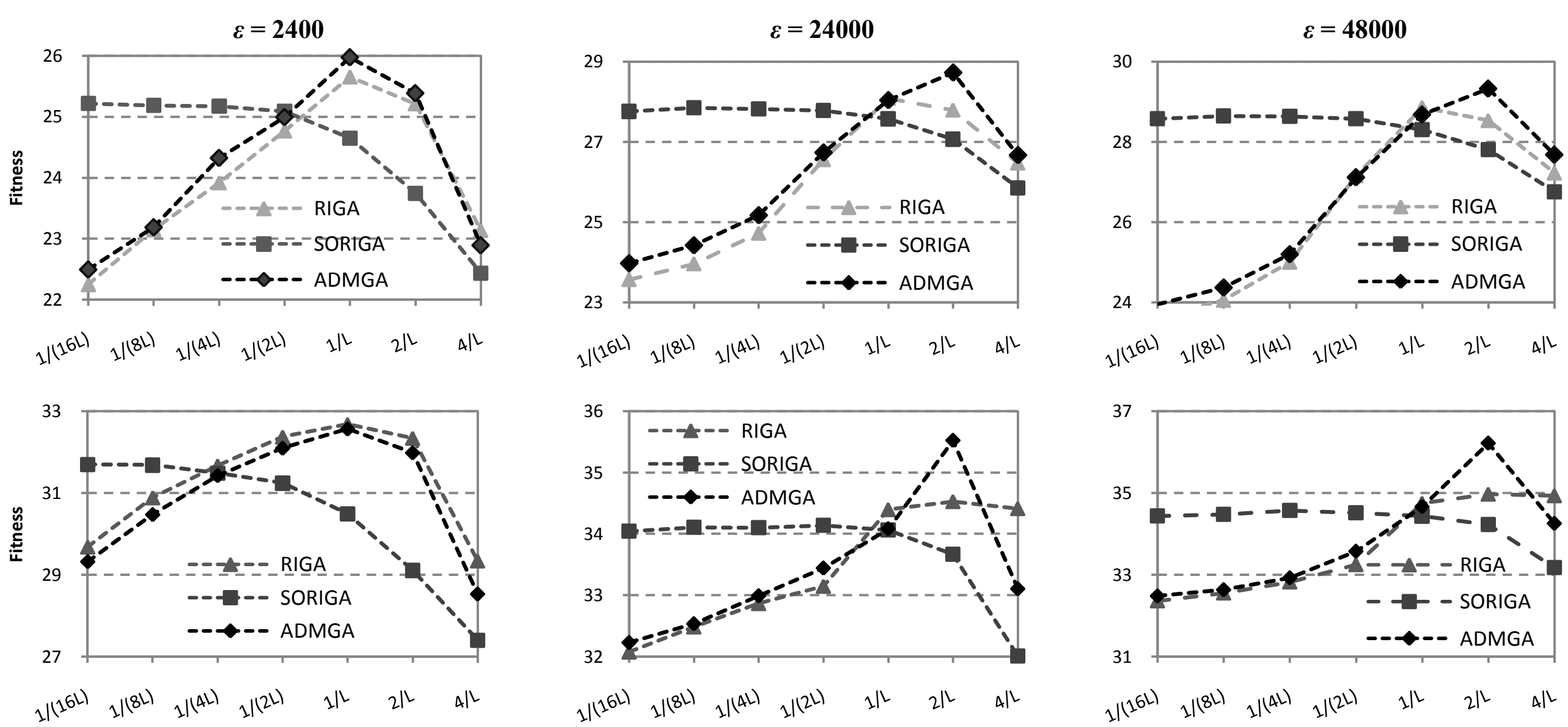

**Fig 9**. Order-3 (first row) and order-4 (second row) dynamic problems. Comparing ADMGA, RIGA and SORIGA with $N$ = 30. RIGA and SORIGA's parameter $r_r$ is set to 4.

**Table 2**. Statistical comparison of algorithms by paired two-tailed t-tests with 58 degrees of freedom at a 0.05 level significance. The t-test results regarding ADMGA and RIGA or SORIGA are shown as + signs when ADMGA is significantly better, - signs when ADMGA is significantly worst, and ~ signs when the algorithms are statistically equivalent. RIGA (3-order) with $p_m = 1/L$ . ADMGA and RIGA (4-order) with $p_m = 1/L$ when ε = 2400 and $p_m = 2/L$ when ε = 24000 and ε = 48000. SORIGA with $p_m = 1/(8\times L)$ for all ε values.

| | | ε = 2400 | | | | ε = 24000 | | | | ε = 48000 | | | |
|---|---|---|---|---|---|---|---|---|---|---|---|---|---|
| | | ρ=0.05 | ρ=0.3 | ρ = 0.6 | ρ = 0.95 | ρ = 0.05 | ρ = 0.3 | ρ = 0.6 | ρ = 0.95 | ρ = 0.05 | ρ = 0.3 | ρ = 0.6 | ρ = 0.95 |
| order-3 | RIGA | + | + | ~ | ~ | + | + | + | + | ~ | + | + | + |
| | SORIGA | + | + | ~ | + | + | + | + | + | + | + | + | + |
| order-4 | RIGA | ~ | ~ | - | - | + | + | + | ~ | + | + | + | + |
| | SORIGA | + | + | + | + | + | + | + | + | + | + | + | - |

### *6.3. Assortative and Dissortative Mating GAs*

In this subsection, ADMGA is compared with the previously proposed negative and positive AMGA [24] [55]. In addition, the dissortative and assortative mating effects on the behaviour of a GGA are also investigated. Since AMGA's are GGAs with a mating restriction, from now on we will refer to the former as AMGGA (the only difference between both algorithms is the way the second parent is select, as described in section 3).

Figure 10 shows the result attained by nAMGGA (dissortative) and pAMGGA (assortative) on order-3 problems and compares them to GGA. Two important conclusions can be drawn. First, nAMGGA is able to outperform GGA in the three types of scenarios. Second, the assortative mating scheme decreases the performance of GGA. (The same results arise on order-4 problems.) Dissortative mating seems to be very effective on this type of dynamic problems, since both ADMGA and nAMGGA are able to outperform GGA on a wide range of scenarios. The structure of the problems is a possible explanation for these results. Since order-4 problems are deceptive (and order-3 are quasi-deceptive) GAs cannot make use of their abilities to combine lower order building blocks to form higher order. Instead, the algorithms must explore the search space, creating radically novel solutions from the ones present in the population. Dissortative mating, by enforcing crossover between dissimilar individuals, enables the GA with the exploratory character needed to tackle deceptive problems. The same character helps the dissortative GAs to react to changes by maintaining and regaining population diversity after a change.

However, nAMGGA has one drawback: the pool size $n$ needs to be tuned. Figure 11 shows how the mean best-of-generation fitness values vary when the pool size changes (please note that GGA is a nAMGGA with $n = 0$). The graphics illustrate the referred difficulties. While with the $N = 30$ configuration (first row) $n = 4$ is the best choice for the pool size, when $N$ is increased to 60 the results are not so clear. Doubling the population size was expected to change the optimal pool size, but it was also expected that it would increase it, in order to maintain a similar level of pressure on the mating between dissimilar individuals. However, results (second row of figure 11) are the opposite: lower $n$ now performs better. This simple test illustrates how hard it can be to hand-tune nAMGGA's parameter $n$. It is also a possible explanation for the reports in [31] to conclude that nAMGGA decreased GGA performance on 4-trap functions.

Another interesting outcome observed in figures 10 and 11 is how the optimal $p_m$ values vary with the different strategies and intensity of the restrictions. Figure 10 shows that the optimal value decreases when moving from an assortative to a dissortative strategy, passing through an intermediate value with the random mating GA (GGA). These results confirm the assumptions in [56] and the results in [55]. In addition, figure 11 shows that when the intensity of dissortative mating is increased, the optimal $p_m$ also decreases.

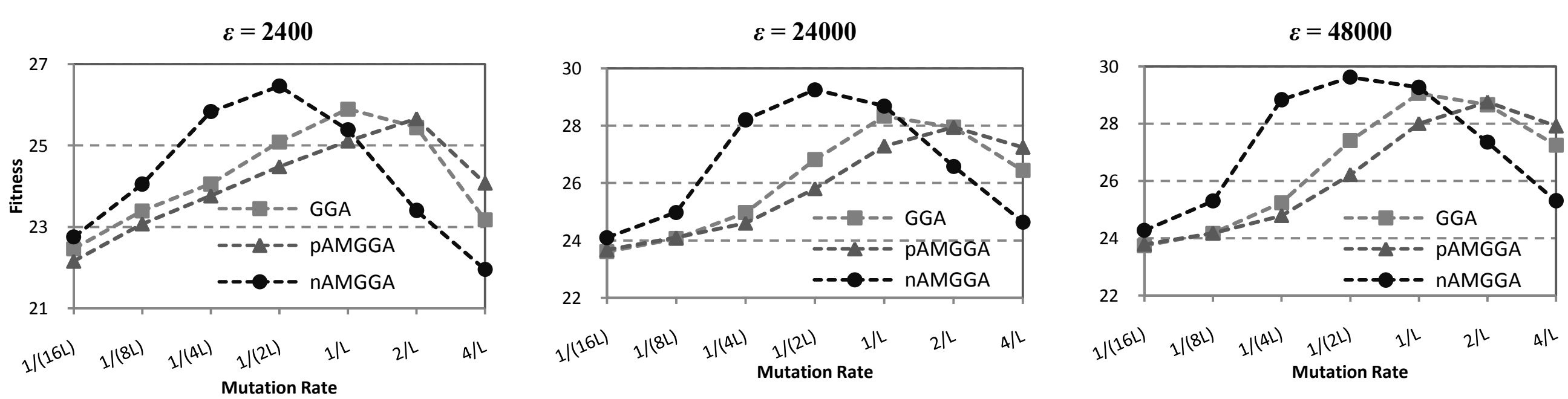

**Fig. 10**. Order-3 dynamic problems. A comparison of GGA and GGA with dissortative (nAMGGA) and assortative mating (pAMGGA). Population size $N = 30$ and pool size $n = 4$ (AMGGAs).

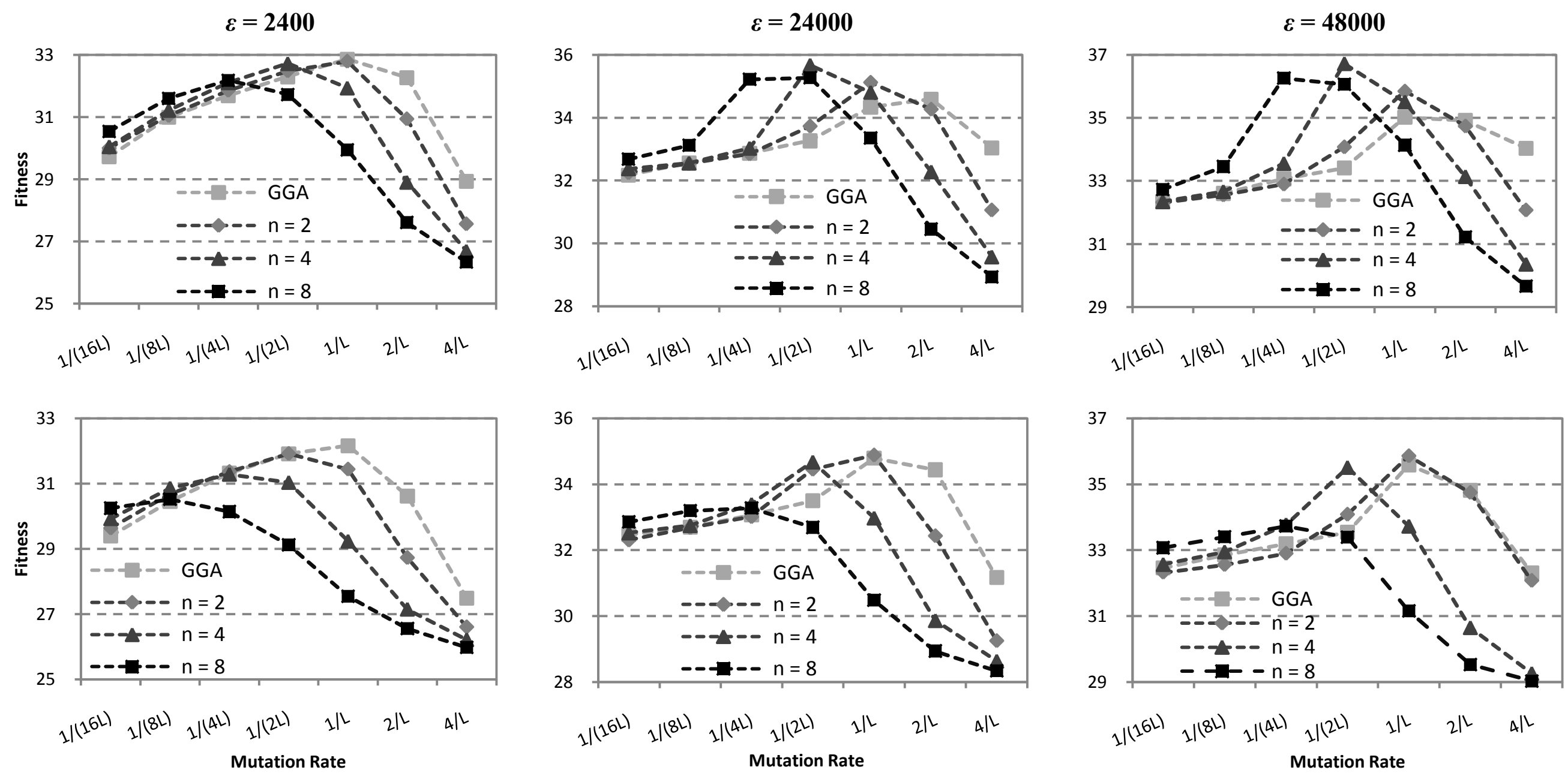

**Fig 11.** Order-4 dynamic problems. Effects of increasing pool size on nAMGGA's performance. $N = 30$ (top row) and $N = 60$ (b row).

A first comparison between nAMGGA and ADMGA was carried out by selecting the best configurations and submitting the mean best-of-generation values to statistical tests. The results are described in table 3 and show that nAMGGA − when properly tuned − is capable of attaining better or at least equivalent results in most of the scenarios. However, its efficiency when compared to ADMGA decreases when the order of the problem is raised. According to these results, it may be asked what happens if the DOP gets even harder.

### *6.4. 5-order dynamic problems*

In order to investigate the behaviour of ADMGA and nAMGGA on 5-order DOPs a first test was conducted with the exact same settings ($\varepsilon$, $\rho$) as in the previous subsections. Again, small populations ($N = 30$ and $N = 60$) performed better than large ones ($N = 120$, $N = 240$ and $N = 480$ were also tested). Figure 12 shows the mean best-of-generation values attained by the best configurations of the algorithm on slow environments (ε = 48000). Unlike the previous problems – see table 3 – ADMGA is now capable of outperforming or at attain similar results as nAMGGA. ADMGA is becoming more effective as the order of the problem increases. However, by looking at the fitness values and observing some best fitness plots – examples are shown in figure 13 −, it is also clear that the algorithms are converging to local optima in most of the runs and periods of change. The problem is that such small populations do not hold the diversity needed to solve this hard instance of the trap functions and so the tests are comparing poor results. It is thus necessary to design a different experiment in order evaluate the differences between the GAs.

**Table 3.** Statistical comparison of algorithms by paired two-tailed t-tests with 58 degrees of freedom at a 0.05 level significance. The t-test result regarding ADMGA and nAMGGA is shown as + sign when ADMGA is significantly better, - sign when ADMGA is significantly worst, and ~ sign when the algorithms are statistically equivalent. ADMGA with $p_m = 1/L$ when $\varepsilon = 2400$ and $p_m = 2/L$ when $\varepsilon = 24000$ and ε = 48000. nAMGGA with $p_m = 1/(L)$ for all $\varepsilon$ values when n = 2. nAMGGA with $p_m = 1/(2\times L)$ for all $\varepsilon$ values when n = 4. nAMGGA with $p_m = 1/(4L)$ for all $\varepsilon$ values when n = 8.

| | | | ε = 2400 | | | | ε = 24000 | | | | ε = 48000 | | | |
|---|---|---|---|---|---|---|---|---|---|---|---|---|---|---|
| | | $n$ | ρ=0.05 | ρ=0.3 | ρ = 0.6 | ρ = 0.95 | ρ = 0.05 | ρ = 0.3 | ρ = 0.6 | ρ = 0.95 | ρ = 0.05 | ρ = 0.3 | ρ = 0.6 | ρ = 0.95 |
| order-3 | nAMGGA | 4 | - | - | - | ~ | ~ | - | - | - | ~ | ~ | - | - |
| order-4 | nAMGGA | 2 | ~ | - | - | ~ | + | + | + | ~ | ~ | + | + | ~ |
| | nAMGGA | 4 | ~ | ~ | - | ~ | ~ | - | + | ~ | - | - | ~ | ~ |
| | nAMGGA | 8 | ~ | + | + | ~ | ~ | + | + | ~ | - | ~ | + | ~ |

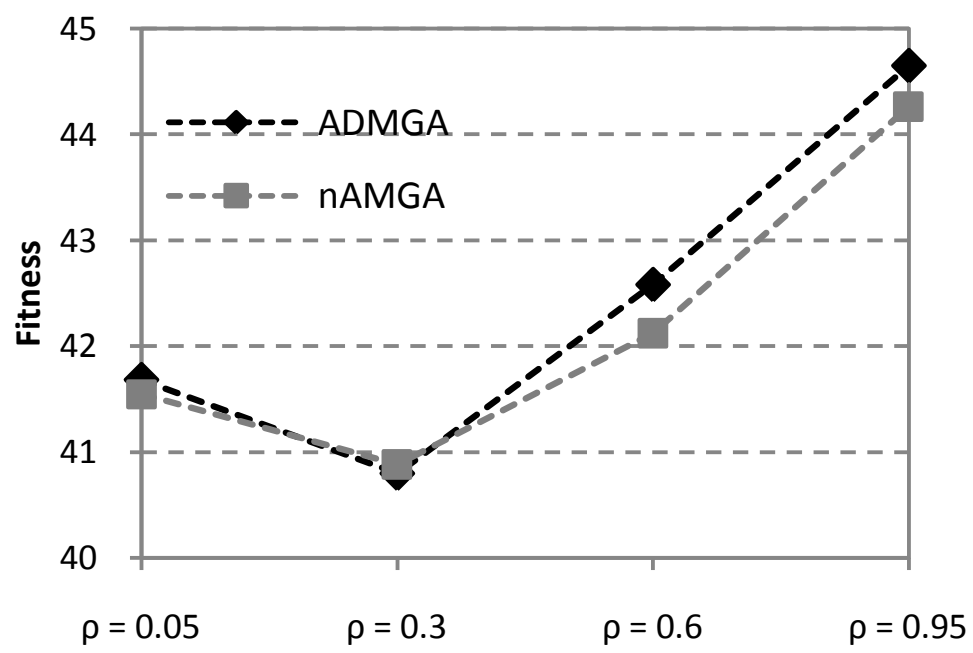

**Fig. 12.** 5-order dynamic problems. ADMGA and nAMGGA ($n$ = 4) mean best-of-generation fitness values on the four scenarios with $\varepsilon$ = 48000. $N$ = 30.

For that purpose, $\varepsilon$ values were increased to 10000, 100000 and 200000, while $\rho$ values remained the same. This way, it is possible to enlarge population size up to 1000 individuals (the fastest scenario still allows 10 generations between each change). Obviously, nAMGGA's optimal pool size is not expected to be the same. Therefore, other values were tested as shown in figure 14, where it is clear that ADMGA is capable of outperforming nAMGGA when speed is slow. Table 4 shows the statistical tests. nAMGGA still performs better on fast environments, but as for slower functions, ADMGA is now better than nAMGGA (except for the scenarios with ρ = 0.95 and for the same reasons exposed in sub-section 5.1).

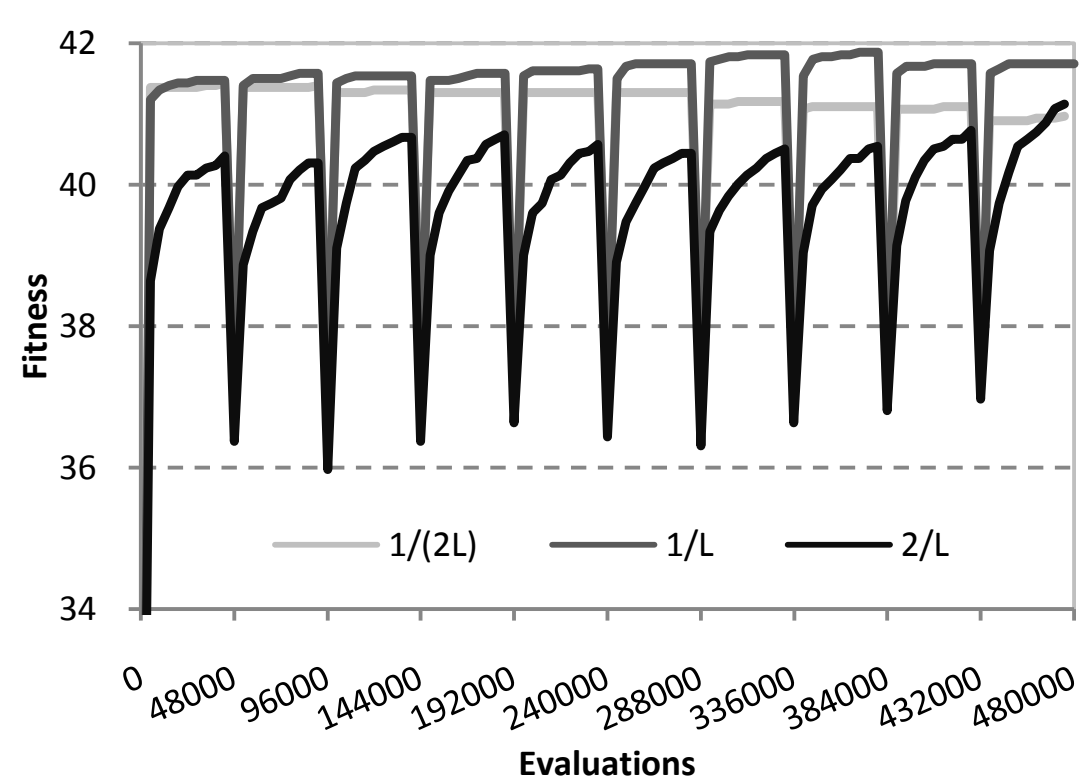

**Fig. 13.** 5-order problem with $\varepsilon$ = 48000 and $\rho$ = 0.05. nAMGGA best fitness values throughout the run with different $p_m$ values. The algorithm is performing very badly; it gets stuck very close (and very soon) to the local (deceptive) optima with fitness 40.

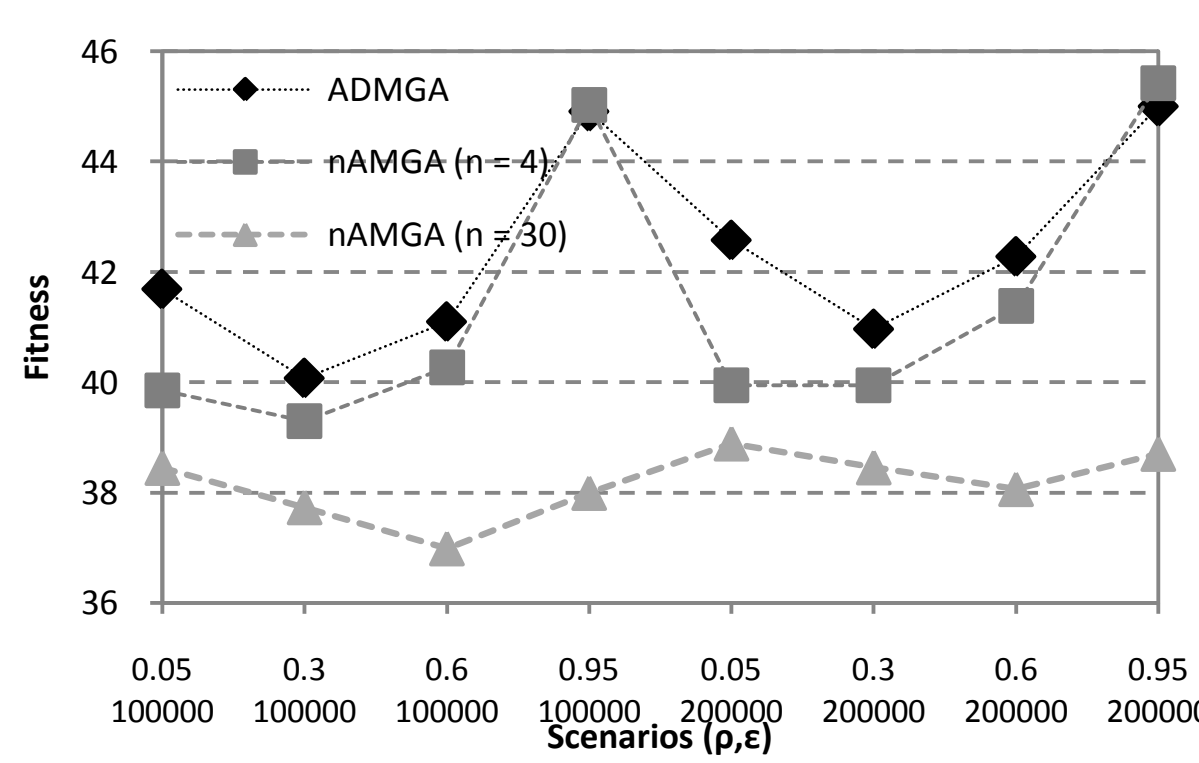

**Fig. 14.** Mean best-of-generation values attained by ADMGA (pm = 1/L), nAMGGA with n = 4 (pm = 1/(8L) ) and nAMGA with n = 30 (pm = 1/(16L) ) on eight 5-order dynamic scenarios. nAMGGA's pool size is set to 4. $N$ = 1000.

**Table 4.** Statistical tests. The + sign means that ADMGA is significantly better, - sign means that ADMGA is significantly worst, and ~ sign when the algorithms are statistically equivalent. ADMGA with $p_m = 1/L$. nAMGGA (n = 4) with $p_m = 1/(8\times L)$

| | | ρ=0.05 | ρ=0.3 | ρ=0.6 | ρ=0.95 |
|---|---|---|---|---|---|
| nAMGGA | ε = 10000 | - | - | - | - |
| | ε = 100000 | + | + | + | ~ |
| | ε = 200000 | + | + | + | ~ |

## 7. Conclusions and Future Work

This paper presents a study on dissortative mating and dynamic optimization and investigates how the *Adaptive Dissortative Mating Genetic Algorithm* (ADMGA) behaves on dynamic deceptive problems. Experimental results show that ADMGA can improve Genetic Algorithms' performance on dynamic trap problems when considering a wide range of possible scenarios. When facing each defined scenario, ADMGA is more capable to track the optima when speed of change is low, which means that it holds abilities to prevent or escape converging populations. ADMGA also outperformed two Random Immigrants Genetic Algorithms (RIGA): a traditional one and another that makes use of a Self-Organized Criticality model to try to overcome some of traditional RIGA's drawbacks. When compared with a previously proposed negative assortative mating GA (nAMGGA), ADMGA outperformed it when solving the hardest instances of the problems. Actually, ADMGA abilities to solve dynamic deceptive functions increased − when compared to nAMGGA − with the order of the problems. In general, the dissortative schemes were more able to tackle deceptive environments, probably because the structure of the problems can make use of the algorithms' ability in exploring the search space and maintaining genetic diversity. Another positive feature of ADMGA is the way it self-regulates the intensity of mating restrictions, unlike nAMGGA, which introduces a new parameter that needs to be hand tuned in order to guarantee a good performance. In addition, ADMGA mating scheme may be extended to other evolutionary algorithms, as long as a metrics for defining the distance between genotypes is defined.

The reasons why dissortative mating schemes behave better in dynamic trap functions may reside on their mating restriction, which forces crossover between dissimilar individuals, thus maintaining the genetic diversity at a higher level. Dynamic optimization requires from evolutionary algorithms a careful balance between exploration and exploitation, and a higher diversity, in general, than required by stationary problems. Besides, deceptive functions can make use of the dissortative mating capabilities to create radically novel solutions, thus increasing exploration and supplying the algorithms with the capacity of escaping local optima. ADMGA, by incorporating a dissortative mating scheme, arises as a good candidate to solve dynamic deceptive functions, and the results confirm the predictions. In addition, by self-regulating the intensity of dissortative mating, ADMGA is provided with means to deal with the different stages of the search, and to react to changes without external actions

Further research will be focused on optimal population size and scalability issues on dynamic environments. A self-adjusting scheme including dissortative and assortative mating is also under consideration. Finally, the design of randomly changing environments could lead to more conclusions on the robustness of the model.

## 8. Acknowledgments

The first author wishes to thank FCT, Ministério da Ciência e Tecnologia, his Research Fellowship SFRH/BD/18868/2004, also partially supported by FCT (ISR/IST plurianual funding) through the POS Conhecimento Program that includes FEDER funds. This work has also been supported by NOHNES project from the Spanish Ministry of Science and Education (TIN2007-68083-C02-01). The authors also wish to thank R. Tinós and S. Yang for providing SORIGA's source code.